%% file: arXiv_version_230407.tex
\documentclass[mnsc,nonblindrev]{informs3} 

\OneAndAHalfSpacedXI 
\usepackage{hyperref}
\usepackage{bbm}
\usepackage{xfakebold}

\usepackage{endnotes}
\let\footnote=\endnote

\usepackage{natbib}
 \bibpunct[, ]{(}{)}{,}{a}{}{,}%
 \def\bibfont{\small}%
 \def\BIBand{and}%

\usepackage{algorithm}
\usepackage{algorithmic}
\usepackage{subfigure}
\usepackage{graphicx}
\usepackage{makecell}

\usepackage{ulem}
\usepackage{booktabs}
\usepackage{multirow}
\usepackage{comment}

\usepackage{hyperref}

\usepackage{xcolor}
\hypersetup{
    colorlinks,
    linkcolor={red!50!black},
    citecolor={blue!50!black},
    urlcolor={blue!80!black}
}
\TheoremsNumberedThrough     
\ECRepeatTheorems

\EquationsNumberedThrough    

\newenvironment{assumption'}[1]
  {\renewcommand{\theassumption}{\arabic{assumption}$'$}%
   \addtocounter{assumption}{-1}%
   \begin{assumption}}
  {\end{assumption}}

\newcommand{\fbseries}{\unskip\setBold\aftergroup\unsetBold\aftergroup\ignorespaces}
\usepackage{amsmath}
\usepackage{amsfonts}
\usepackage{amssymb}
\usepackage{dsfont}

\begin{document}



\RUNTITLE{Best Arm Identification with Fairness Constraints on Subpopulations}

\TITLE{Best Arm Identification with Fairness \\Constraints on Subpopulations\footnote{We thank Candace Yano for helpful discussions. All errors are ours. Email: \url{wuyh@berkeley.edu},  \url{zyzheng@berkeley.edu}, \url{tingyu_zhu@berkeley.edu}.}}




\ARTICLEAUTHORS{%
\AUTHOR{Yuhang Wu, Zeyu Zheng, Tingyu Zhu}
\AFF{ Department of Industrial Engineering and Operations Research, University of California, Berkeley, CA}
 }

\ABSTRACT{%
We formulate, analyze and solve the problem of best arm identification with fairness constraints on subpopulations (BAICS). Standard best arm identification problems aim at selecting an arm that has the largest expected reward where the expectation is taken over the entire population. The BAICS problem requires that an selected arm must be fair to all subpopulations (e.g., different ethnic groups, age groups, or customer types) by satisfying constraints that the  expected reward conditional on every subpopulation needs to be larger than some thresholds. The BAICS problem aims at correctly identify, with high confidence, the arm with the largest expected reward from all arms that satisfy subpopulation constraints. We analyze the complexity of the BAICS problem by proving a best achievable lower bound on the sample complexity with closed-form representation. We then design an algorithm and prove that the algorithm's sample complexity  matches with the lower bound in terms of order. 
 A brief account of numerical experiments are conducted to illustrate the theoretical findings. 
}%


%

\maketitle

%

\input{full}
 
\bibliographystyle{informs2014}


\input{arXiv_version_230407.bbl}
\end{document}

%% file: full.tex
\section{Introduction}\label{sec:intro}

Many decision making problems naturally give rise to setting where there are a number of different policies (or systems, designs) each with unknown expected performances, from which the decision maker wants to select the policy with the best expected performance. Even though the expected performances are unknown, the decision maker generally has access to observe independent noisy samples of the expected performance for each policy. The statistically principled way of identifying the best policy through the noisy samples has been a fundamental research problem in several research areas. Some early statistical work includes \cite{bechhofer1954single} and \cite{bechhofer1995design}. In the stochastic simulation literature, the research problem is called \textit{ranking and selection} (R\&S); see \cite{hong2021review}, \cite{hunter2017parallel}, \cite{chick2006subjective} and \cite{kim2006selecting} for reviews. In the multi-armed bandit literature, the research problem is called \textit{best arm identification} (BAI); see \cite{audibert2010best}, \cite{garivier2016optimal}, \cite{kaufmann2016complexity}, fore references. \cite{ma2017efficient} and \cite{glynn2015selecting} have discussed some connections between the two literature. The R\&S literature and BAI literature differ in assumptions and analysis tools. Our work is positioned in both literature, and adopts the assumptions and analysis tools in the BAI literature. 

In this work, we consider the problem of Best Arm Identification with fairness Constraints on Subpopulations (BAICS). We briefly discuss the problem setting of BAICS and the meaning of fairness constraints on subpopulations. The formal setting with precise mathematical formulation is introduced in Section \ref{sec:setting}. In BAICS, each arm represents a policy in consideration to be used on an entire population. There are in total multiple different arms in competition. The expected reward of an arm is typically measured on the entire population. The classical BAI problem, for example, aims at identifying the arm with the largest expected reward over the entire population. For some applications, the population consists of several subpopulations. For example, a subpopulation may represent a ethnic group defined through different cultural background; a subpopulation may also represent a subpopulation of customers defined through different consumption needs. Fairness constraints on subpopulations refer to that the expected reward of an arm conditional on any subpopulation cannot be lower than some pre-specified thresholds. The implication is that the constraints require a policy to be ``fair" to all subpopulations and is not allowed to ``sacrifice" any subpopulation. For example, the constraints can be that the expected reward of an arm conditional on every subpopulation cannot be lower than zero. Given the constraints on subpopulations, the set of arms are classified into two subsets: feasible (satisfying the constraints) and infeasbile (not satisfying the constraints). The BAICS problem aims at selecting an arm that has the largest expected reward among all the feasible arms. 

The BAICS problem and its formulation has direct practical relevance when the decision maker not only cares about maximizing expected rewards, but also cares about each subpopulation's benefit. In particular, the presence of fairness constraints prohibits a decision maker to improve expected reward over the entire population by implicitly exploiting or hurting some subpopulation, which would be unfair to them. The BAICS problem formulation also has relevance to the online controlled experiments (A/B tests) with multiple treatments, where the goal is to select the best treatment over all treatments that provide Pareto improvement to all subpopulations. Despite the relevance of the BAICS problem, the algorithms designed for the classical BAI problem may fail or become not effective on the BAICS problem because they do not consider the fairness constraints on subpopulations. 

We make the following contributions in this work.

\begin{itemize}
    \item To our best of knowledge, we are the first to consider the fairness constraints of subpopulations in the context of the best arm identification problem with fixed confidence criterion, and we propose a new formulation called Best Arm Identification with fairness Constraints on Subpopulations (BAICS), which incorporates subpopulation fairness constraints into the arm selection process.
 
    \item We derive the asymptotic lower bound on the expected stopping time for all algorithms that are guaranteed to solve the problem with a given confidence level. Such lower bounds provides the best achievable sample complexity order for any algorithm that tackles BAICS. We present an explicit formula along with an intuitive interpretation of the sample complexity.

    \item We design an algorithm that is capable of serving two goals --- to identify the best arm and to ensure that it satisfies all subpopulation constraints. We provide theoretical results to show that it achieves the asymptotically optimal sample complexity lower bound. We compare our algorithm with two other methods and illustrate its efficiency through numerical experiments.
\end{itemize}

  The theoretical tools that we develop in this work to analyze lower and upper bounds on the expected sample complexity are partially inspired by the analysis framework proposed in \cite{garivier2016optimal} to address the standard BAI problem. To the best of our knowledge, there are no works in the R\&S literature and in the BAI literature that specifically considers constraints on the arm/policy performances on each subpopulation. A related but different stream of work is constrained R\&S; see \cite{andradottir2010fully}, \cite{healey2014selection}, \cite{hong2015chance} for example. They generally consider the problem of finding a system with the best primary performance metric under constraints on some secondary performance metric. They do not consider the sampling strategy related to each subpopulation to explore each of the subpopulation constraints.  Another difference is that their analysis framework and tools do not focus on developing matching lower and upper bounds for the sampling complexity.


\section{Setting and Formulation}\label{sec:setting}
The mathematical formulation of  Best Arm Identification with fairness Constraints on Subpopulations (BAICS) is given as follows. Suppose we are given the number of arms $K\geq 2$ and the number of subpopulations $L$. We are also given a vector $\mathbf{q}=(q_1,\cdots,q_L)\in \mathbb{R}^L$ representing the importance of the subpopulations. A typical choice in practice is to take $q_l$ as the proportion of subpopulation $l$ in the total population for $1\leq l\leq L$. We further make the stochastic assumption that observations from arm $k$ and subpopulation $l$ are i.i.d. random variables drawn from Gaussian distributions with some known variance, and the variances are the same for all $k$ and $l$.  Without loss of generality we assume the variance is $1$, so observation of arm $k$ and subpopulation $l$ is given by a normal distribution $P_{\mu_{k,l}} \sim \mathcal{N}(\mu_{k,l},1)$ for $k\in[K]$ and $l\in [L]$, here $[K]=\{1,\cdots,K\}$ and $[L]=\{1,\cdots,L\}$.  Such distribution assumptions are commonly seen in the best arm identification literature, e.g., \cite{shang2020fixed,barrier2022non}. The assumption may also be viewed as a special case of the exponential family distribution assumption with one unknown parameter (mean). The quality, or expected performance, of arm $k$ is $\mu_k = \sum_{l=1} q_l \mu_{k,l}$, which is the weighted average of the means of the arm in different subpopulations. 

A standard best arm identification problem tends to find the arm $k_\text{BAI}$ with the maximum quality, i.e.,
\[
k_\text{BAI}=\arg\max_{k\in[K]}\mu_{k}.
\]
However, arm $k_\text{BAI}$ may perform bad on some subpopulations. As we discussed in Section \ref{sec:intro}, we hope to find an arm $k^*$ such that it not only has a good quality, which means $\Bar{\mu}_{k^*}$ is large, but it also works well on some given subpopulations. Mathematically, we introduce the definition \textit{feasible arm} as follows: an arm $k$ is called a feasible arm, if and only if it is in a \textit{feasible set} $C$:
\begin{equation}\label{def:feasible}
    k\in {C}= \{k\in[K]\big\vert \mu_{k,m}\ge 0,\forall m\in[M]\}
\end{equation}
with some known $M$. Intuitively, a feasible arm means it performs not too bad on subpopulations $1$ to $M$. Now, our goal is to find the best arm $k^*$ in the feasible set $C$:
\begin{equation}\label{def:goal}
    k^*=\arg\max_{k\in{C}}\mu_{k}.
\end{equation}
Here, when $C=\emptyset$, we define $k^*=0$. Throughout the paper, we will also say the best feasible arm is $0$ if $C=\emptyset$.

At each step $t$, the algorithm selects an arm $A_t\in [K]$ and a subpopulation $I_t\in [L]$ based on previous choices and outcomes. After that an observation $X_t$ is obtained, which is a sample drawn from $P_{\mu_{A_t,I_t}}$. This naturally defines a filtration generated by all information up to step $t$ denoted $\mathcal{F}_t = \sigma(\{I_s,A_s,X_s\}_{s=1,2,\cdots,t})$. The algorithm then chooses $A_{t+1},I_{t+1}$ which is $\mathcal{F}_t$-measurable. We further define $N_{a,i}(t) =\sum_{s=1}^{t} \mathbbm{1}(I_s = i, A_s = a)$ and $N_{a}(t)=\sum_{s=1}^{t} \mathbbm{1}(A_s = a)$.

In addition, since the distributions $\left(P_{\mu_{k,l}}\right)_{k\in[K],l\in[L]}$ are assumed to be Gaussian in $\mathcal{P}=\{P\big\vert P\sim \mathcal{N}(\mu,1)\}$, we may hence identify any bandit instance with its matrix of means ${\fbseries \mu} \in\mathbb{R}^{K\times L}$. Simple calculation shows that the Kullback-Leibler divergence between two Guassian distribution $P\sim \mathcal{N}(\mu,1)$ and $Q\sim\mathcal{N}(\nu,1)$ is given by $\mathbf{KL}(P,Q)=\frac{1}{2}(\mu-\nu)^2.$ 

 Denote by $\mathcal{S}$ a set of Gaussian bandit models such that, each bandit model ${\fbseries\mu}$ in $\mathcal{S}$ satisfies $(1)$ either has a unique optimal feasible arm with all constraints strictly satisfied; $(2)$ Or all arms of it have at least one subpopulation constraint strictly violated. That is, for each ${\fbseries\mu}\in\mathcal{S}$, denote the feasible set by $C({\fbseries\mu})$, then $(1)$ either there exists an arm $k^*({\fbseries\mu})\in C({\fbseries\mu})$ such that $\mu_{k^*({\fbseries\mu})}>\max_{k\in C}\{\mu_k\big\vert k\neq k^*({\fbseries\mu})\}$ and $\mu_{k^*({\fbseries\mu}),l}>0$ for $l\in[M]$; $(2)$ Or $C({\fbseries\mu})=\emptyset$ and $\forall k\in[K]$ there exists $l\in[M]$ such that $\mu_{k,l}<0$. In the latter case, we define $k^*({\fbseries\mu})=0$, then $k^*({\fbseries\mu})$ is always unique for ${\fbseries\mu}\in \mathcal{S}$. The definition of $\mathcal{S}$, though seems complicated at first glance, is just to make sure the bandit model has some ``gap" so the best arm is unique and can be identified. We will focus on bandit models in $\mathcal{S}$ thereafter.
 
 In this paper, we focus on the \textit{fixed-confidence setting} with risk level $\delta$. An algorithm under this setting is called $\delta$-PAC if it gives a stopping time $\tau_\delta$ with respect to $\mathcal{F}_t$, a $\mathcal{F}_{\tau_\delta}$-measurable recommendation $\hat{k}_{\tau_\delta}\in \{0\}\cup [K]$, and 
\begin{equation}\label{def:delta_PAC}
\begin{aligned}
       & \forall {\fbseries\mu}\in\mathcal{S}, \mathbb{P}_ {\fbseries\mu}(\tau_\delta<+\infty)=1,\\
       & \mathbb{P}_ {\fbseries\mu}(\hat{k}_{\tau_\delta}\neq k^*({\fbseries\mu}) )\leq \delta.
\end{aligned}
\end{equation}

\section{Lower Bounds on the Sample Complexity}\label{sec:lower_bound}

In this section, we prove and analyze lower bounds on the sample complexity of $\delta$-PAC algorithms for the BAICS problem. The lower bounds represent the best achievable sample complexity for any algorithm that can return the correct solution with $\delta$-PAC guarantee. Through the lower bounds we will be able to characterize how complex the problem at least is.

\subsection{General sample complexity of best arm identification}\label{subsec:general_lower_bound}
First, we introduce 
\begin{equation}\label{def:alt}
    \text{Alt}({\fbseries\mu}):=\{{\fbseries\lambda}\in \mathcal{S}\big\vert k^*({\fbseries\mu})\neq k^*({\fbseries\lambda})\},
\end{equation}
the set of problems where the optimal feasible arm is not the same as in ${\fbseries\mu}$, and $\Sigma_{K\times J}=\{\mathbf{w}\in \left(\mathbb{R}_+\cup \{0\}\right)^{KL}\big\vert w_1+\cdots+w_{KJ}=1\}$ the set of probability distributions on $[K]\times[L]$. Then we have following lower bound for the sample complexity. 

\begin{theorem}\label{thm:lower_bound}
Let $\delta\in(0,1)$ and $\mathbf{q}\in\mathbb{R}^J$. For any $\delta$-PAC policy and any bandit model ${\fbseries\mu}\in\mathcal{S}$, 
\begin{equation}\label{eq:lower_bound}
    \mathbb{E}_{{\fbseries\mu}}[\tau_\delta]\geq T^*({\fbseries\mu})\text{kl}(\delta,1-\delta) \text{ and } \liminf_{\delta\to 0}\frac{\mathbb{E}_{{\fbseries\mu}}[\tau_\delta]}{\ln(1/\delta)}\geq T^*(\mu),
\end{equation}
where
\begin{equation}\label{eq:T_star}
    \begin{aligned}
        T^*({\fbseries\mu})^{-1}&=\frac{1}{2}\sup_{\mathbf{w}\in \Sigma_{K\times J}}\inf_{{\fbseries\lambda}\in \text{Alt}({\fbseries\mu})}\sum_{k\in[K]}\sum_{l\in[L]}w_{k,l}\left(\mu_{k,l}-\lambda_{k,l}\right)^2,\\
    \end{aligned}
\end{equation}
and $\text{kl}(\delta,1-\delta)$ is the KL divergence of two Bernoulli distributions of parameter $\delta$ and $1-\delta$.
\end{theorem}

The proof of Theorem \ref{thm:lower_bound} can be directly adapted from Theorem 1 of \cite{russac2021b} by noting that $\textbf{KL}(P_{\mu_{k,l}},P_{\lambda_{k,l}})=\frac{(\mu_{k,l}-\lambda_{k,l})^2}{2}$. This general lower bound origins from \cite{kaufmann2016complexity} and \cite{garivier2016optimal}. Here, $T^*({\fbseries\mu})$ characterizes the difficulty of the problem. $\mathbf{w}^*$ that achieves the supreme of \eqref{eq:T_star} can be intuitively understood as the optimal sampling proportions of total samples for each arms and subpopulations. We will also see in next subsection that, although $T^*({\fbseries\mu})$ has a similar form as it appears in \cite{kaufmann2016complexity} and \cite{garivier2016optimal}, it is essentially different from that in BAI because of the different structure of $\text{Alt}({\fbseries\mu})$ in BAICS problem.

\noindent\textbf{Remark 1 } It is worth noting that the RHS of \eqref{eq:T_star} is positive (i.e., not zero) when we  consider bandit models in $\mathcal{S}$, so $T^*({\fbseries\mu})$ is well-defined, and this fact will be more clear when we simplify the expression of $T^*({\fbseries\mu})^{-1}$ in Theorem \ref{thm:trade_off}. This emphasizes the justification for focusing on bandit models in $\mathcal{S}$. The intuition is that, for any bandit model ${\fbseries\mu}\notin \mathcal{S}$, we can always construct an alternative ${\fbseries\lambda}\in\text{Alt}({\fbseries\mu})$ as close to it as we want, so the infimum in \eqref{eq:T_star} becomes $0$ and the difficulty of the problem becomes infinity.

\subsection{Implicit tradeoff in BAICS problem}
\label{subsec:trade_off_BAICS}
While Theorem \ref{thm:lower_bound} holds for general BAI problem, we now focus specifically on our BAICS problem and demonstrate that a tradeoff in sampling strategy arises naturally in BAICS. This tradeoff makes the BAICS problem inherently different from BAI, as we will show in detail. In this subsection, to provide an intuition for this tradeoff, we will begin with a simple example. After that, we will present a mathematical formulation to further illustrate this relationship.

\noindent\textbf{Example 1: }
Suppose we set $K=3,L=2,M=2$ and $\mathbf{q}=(\frac{1}{2},\frac{1}{2}).$ $\mu_{1,1}=\mu_{1,2}=1$; $\mu_{2,1}=4, \mu_{2,2}=-\varepsilon$; and $\mu_{3,1}=1$, $\mu_{3,2}=1-\varepsilon$, with some $0<\varepsilon<1$. Without constraints, it is easy to see that the BAI problem has $k_\text{BAI}=2$, because $\mu_1 = \frac{1}{2} \mu_{1,1} +\frac{1}{2} \mu_{1,2}=1$, $\mu_2=\frac{1}{2}\mu_{1,1}+\frac{1}{2}\mu_{2,2}=2-\frac{\varepsilon}{2}$ and $\mu_2=\frac{1}{2}\mu_{3,1}+\frac{1}{2}\mu_{3,2}=1-\frac{\varepsilon}{2}$. Specifically, since the difference between the means of arms 2 and 1 is $\mu_2-\mu_1=1-\frac{\varepsilon}{2}$, and the difference between the means of arms 2 and 3 is $\mu_2-\mu_3=1$, the gap between the best arm and the other arms is relatively large when $\varepsilon$ is much smaller than 1. However, when we consider subpopulation constraints in the BAICS problem, the best arm is now $k^*=1$. To identify $k^*=1$, we must distinguish that $\mu_{2,2}<0$ because $\mu_2>\mu_1$, and also recognize that $\mu_3<\mu_1$. When $\varepsilon$ goes to $0$, these can be substantially more difficult than BAI because both of above gaps are $\varepsilon$. 

The simple example above highlights the fundamental difference between BAI and BAICS. In the BAI problem, explorations are used to find the arm with the highest mean. However, in the BAICS problem, explorations introduce an implicit tradeoff between optimality and feasibility. In Example 1 with a small $\varepsilon$, to make a conclusion that $k^*=1$, people need to estimate $\mu_{1}$, $\mu_{2,2}$, and $\mu_{3}$ accurate enough, which leads to a natural problem to allocate samples among arm $1$ and arm $3$ for optimality and also subpopulation $2$ of arm $2$ for feasibility. 

We also point out that Example 1 does not mean BAICS is always more difficult than BAI. In fact, if we slightly change the setting to $\mu_{2,2}=-2-\varepsilon,\mu_{3,2}=-1$ and keep others the same as in Example 1, then it is easy to see $k^*=k_\text{BAI}=1$. This time BAICS is easy because we can easily tell $\mu_{2,2},\mu_{3,2}<0$ and arm $1$ is feasible, but BAI is hard because $\mu_1-\mu_2=\frac{\varepsilon}{2}$ is small when $\varepsilon$ is close to $0$. 

 We can gain a deeper understanding of the BAICS problem by consider another variant of Example 1. This time, we change $\mu_{2,1}=1$, $\mu_{3,2}=-1$ and keep others the same. Since now $\mu_1 = 1, \mu_2=\frac{1-\varepsilon}{2}$ and $\mu_3 = 0$, it is again easy to identify $k^*=1$. The interesting thing in this example is that it is not necessary to identify the feasible set $C$ before we find $k^*=1$. Indeed, it is possible that our algorithm can not tell whether $\mu_{2,2}\geq0$ when $\varepsilon$ is small, but it can still recommend $k^*=1$ with risk at most $\delta$ because $\mu_2$ is much smaller than $\mu_1$ and we do not need to know the feasibility of arm $2$. From this example we can see, a naive algorithm that attempts to find the feasible set $C$ before searching for the best arm in $C$ can be inefficient in general, so the BAICS problem is not a straightforward synthesis of finding the feasible set and a standard BAI problem.

Above examples and discussions show that the complexity of a BAICS problem needs not to be related to the corresponding BAI problem without constraints, and the BAICS problem naturally leads to an optimality-feasibility tradeoff and presents unique challenges. We now formally state the theorem that captures this intuition. Notation-wise, when we specify any ${\fbseries\lambda}\in \mathbb{R}^{K\times L}$, we regard $\lambda_{k,l}$ as the $(k,l)$-th element of ${\fbseries\lambda}$ and $\lambda_k = \sum_{l=1}^{L}\lambda_{k,l}$ for $k=1,2,\cdots,K$. 
\begin{theorem}\label{thm:trade_off}
For any ${\fbseries\mu}\in \mathcal{S}$, if $k^*({\fbseries\mu})=0$, 
\begin{equation}\label{eq:k=0}
    T^*({\fbseries\mu})^{-1}=\frac{1}{2}\max_{\mathbf{w}\in\Sigma_{K\times L}}\min_{k\in[K]}\sum_{l\in[M],\mu_{k,l}<0}w_{k,l}\mu_{k,l}^2;
\end{equation}
If  $k^*({\fbseries\mu})\neq0$, without loss of generality we assume $k^*({\fbseries\mu})=1$, then
\begin{equation}\label{eq:k=1}
    T^*({\fbseries\mu})^{-1}=\frac{1}{2}\max_{\mathbf{w}\in\Sigma_{K\times L}}\min\left(f^\text{opt}_{\fbseries\mu}(\mathbf{w}),f^\text{fea}_{\fbseries\mu}(\mathbf{w})\right),
\end{equation}
where
\[
f^\text{opt}_{\fbseries\mu}(\mathbf{w})=\min_{2\leq k\leq K}\min_{\substack{{\fbseries\lambda}\in \mathbb{R}^{K\times L}\\ \lambda_k\geq \lambda_1\\\forall l\in[M],\lambda_{k,l}\geq0}}\left(\sum_{l\in[L]}w_{1,l}(\mu_{1,l}-\lambda_{1,l})^2+\sum_{l\in[L]}w_{k,l}(\mu_{k,l}-\lambda_{k,l})^2\right),
\]
and 
\[
f^\text{fea}_{\fbseries\mu}(\mathbf{w})=\min_{l\in[M]}w_{1,l}\mu_{1,l}^2.
\]
\end{theorem}
The proof of Theorem \ref{thm:trade_off} is given in Section \ref{subsec:proof_thm2}. In Theorem \ref{thm:trade_off}, \eqref{eq:k=0} gives the sample complexity lower bound when there is no feasible arm, i.e. $C({\fbseries\mu})=\emptyset$. As for the case $C({\fbseries\mu})\neq \emptyset$, the complexity of the problem $T^*({\fbseries\mu})$ defined in \eqref{eq:T_star} now consists of two terms $f^\text{opt}_{\fbseries\mu}(\mathbf{w})$ and $f^\text{fea}_{\fbseries\mu}(\mathbf{w})$, which reflect the credibility of optimality and the credibility of feasibility, respectively. Briefly, $f^\text{opt}_{\fbseries\mu}(\mathbf{w})$ can be interpreted as a measure of assurance that other feasible arms are not as good as arm $1$, and $f^\text{fea}_{\fbseries\mu}(\mathbf{w})$ is a measure of assurance that arm $1$ is feasible. The smaller these two values are, the less assurance and therefore the more difficult the problem becomes. The notion ${\mathbf{w}}$ is the proportions of samples for each arm and each subpopulation. $T^*({\fbseries\mu})^{-1}$ is then obtained through maximizing the minimum of $f^\text{opt}_{\fbseries\mu}(\mathbf{w})$ and $f^\text{fea}_{\fbseries\mu}(\mathbf{w})$, which can be interpretted as a tradeoff between minimizing the complexity of optimality and the complexity of feasibility.

\subsection{Proof of Theorem \ref{thm:trade_off}}\label{subsec:proof_thm2}
In this part, we discuss the proof of Theorem \ref{thm:trade_off}. The basic idea is to take a close look at how to fix $\mathbf{w}$ and construct a close-by alternative bandit instance ${\fbseries\lambda}\in \text{Alt}({\fbseries\mu})$. Recall that 
$$T^*({\fbseries\mu})^{-1}=\frac{1}{2}\sup_{\mathbf{w}\in \Sigma_{K\times J}}\inf_{{\fbseries\lambda}\in \text{Alt}({\fbseries\mu})}\sum_{k\in[K]}\sum_{l\in[L]}w_{k,l}\left(\mu_{k,l}-\lambda_{k,l}\right)^2.$$
First we consider the case $C({\fbseries\mu})=\emptyset$, then $k^*({\fbseries\mu})=0$, so $\text{Alt}({\fbseries\mu})=\{{\fbseries\lambda}\in\mathcal{S}\big\vert k^*({\fbseries\lambda})\neq 0\}$. That is, as long as ${\fbseries\lambda}$ has one feasible arm, then it is in the alternative set $\text{Alt}({\fbseries\mu})$. Fix $\mathbf{w}$, for any $i\in[K]$, to make sure $i\in C({\fbseries\mu})$, we only require $\lambda_{i,l}\geq 0$ for all $l\in[M]$, so we only need to set $\lambda_{i,l}=0$ for those $l\in[M]$ such that $\mu_{i,l}<0$, and take other $\lambda_{k,l}$ just to be $\mu_{k,l}$, then $$\inf_{{\fbseries\lambda}\in \text{Alt}({\fbseries\mu})}\sum_{k\in[K]}\sum_{l\in[L]}w_{k,l}\left(\mu_{k,l}-\lambda_{k,l}\right)^2=\min_{k\in[K]}\sum_{l\in[M],\mu_{k,l}<0}w_{k,l}\mu_{k,l}^2.$$ 
It is easy to see this is a continuous function of $\mathbf{w}$ and the domain of $\mathbf{w}$ is compact, so the supremum can be attained by some $\mathbf{w}^*({\fbseries\mu})$, and we obtain \eqref{eq:k=0}. 

As for the case $C({\fbseries\mu})\neq\emptyset$, without loss of generality we assume $k^*({\fbseries\mu})=1$. Again we fix $\mathbf{w}$. To construct an alternative instance ${\fbseries\lambda}\in \text{Alt}({\fbseries\mu})$, we have two different ways. The first option consists in taking an arm $i>1$ and augment means of its subpopulations on the alternative model such that it becomes above arm $1$. Otherwise, it is possible to shrink the mean of one subpopulation of arm $1$ such that it becomes infeasible on the alternative. We will now consider each of them separately.

For the first option, suppose we want to take arm $i>1$ and augment its means. Then in the alternative model ${\fbseries\lambda}$, we would expect  $\lambda_k\geq \lambda_1$ and $\forall l\in[M],\lambda_{i,l}\geq0$, and for other $k\neq1,i$, we take $\lambda_{k,l}=\mu_{k,l}$ to minimize $\sum_{\substack{k\in[K]\\k\neq1,i}}\sum_{l\in[L]}w_{k,l}\left(\mu_{k,l}-\lambda_{k,l}\right)^2$ to be $0$. Note here we only require $\lambda_k\geq \lambda_1$ because we can always add a small number to some $\lambda_{k,l}$ to make the inequality strict. Then in this case, we obtain 
$$f^\text{opt}_{\fbseries\mu}(\mathbf{w})=\min_{2\leq k\leq K}\min_{\substack{{\fbseries\lambda}\in \mathbb{R}^{K\times L}\\ \lambda_k\geq \lambda_1\\\forall l\in[M],\lambda_{k,l}\geq0}}\big(\sum_{l\in[L]}w_{1,l}(\mu_{1,l}-\lambda_{1,l})^2+\sum_{l\in[L]}w_{k,l}(\mu_{k,l}-\lambda_{k,l})^2\big).$$

For the other way, we want to shrink the mean of one subpopulation of arm $1$ to make it infeasible. Thus, we only need to modify $\lambda_{1,l}$ to be $0$ for some $l\in[M]$ and set all other $\lambda_{k,l}=\mu_{k,l}$. Again we only need $\lambda_{1,l}=0$ because we can subtract it by an arbitrarily small number to make it negative. Now by iterating over $l\in[M]$ we can define
$$f^\text{fea}_{\fbseries\mu}(\mathbf{w})=\min_{l\in[M]}w_{1,l}\mu_{1,l}^2.$$

Combine above two cases together and we obtain 
$$\inf_{{\fbseries\lambda}\in \text{Alt}({\fbseries\mu})}\sum_{k\in[K]}\sum_{l\in[L]}w_{k,l}\left(\mu_{k,l}-\lambda_{k,l}\right)^2=\min\left(f^\text{opt}_{\fbseries\mu}(\mathbf{w}),f^\text{fea}_{\fbseries\mu}(\mathbf{w})\right).$$
It is easy to see $\min\left(f^\text{opt}_{\fbseries\mu}(\mathbf{w}),f^\text{fea}_{\fbseries\mu}(\mathbf{w})\right)$ is continuous, so the supremum on a compact set can be replaced by the maximum, then
$$ T^*({\fbseries\mu})^{-1}=\frac{1}{2}\sup_{\mathbf{w}\in \Sigma_{K\times J}}\inf_{{\fbseries\lambda}\in \text{Alt}({\fbseries\mu})}\sum_{k\in[K]}\sum_{l\in[L]}w_{k,l}\left(\mu_{k,l}-\lambda_{k,l}\right)^2=\frac{1}{2}\max_{\mathbf{w}\in\Sigma_{K\times L}}\min\left(f^\text{opt}_{\fbseries\mu}(\mathbf{w}),f^\text{fea}_{\fbseries\mu}(\mathbf{w})\right),$$
which finishes the proof.

\section{Algorithm Design and Complexity Analysis}\label{sec:alg}
In this section, we develop an algorithm to solve the BAICS problem with $\delta$-PAC guarantee. We prove upper bound on the sample complexity of the proposed algorithm. We show that the upper bound matches the proved lower bound in the order. 

To develop our algorithm, we adapt the Track-and-Stop algorithm introduced in  \cite{garivier2016optimal} to the BAICS problem. We first discuss the sampling rule and its calculation. Then we give the stopping rule, our recommendation of the best feasible arm, and the threshold for stopping. Finally, we give the convergence result of our algorithm to show it is asymptotically optimal in the sense that it matches the sample complexity lower bound asymptotically.

\subsection{The sampling rule and its calculation}\label{subsec:sampling_rule}
In this part, we first give a high level overview of the sampling rule and then give the details of our implementation. Suppose we are given the number of arms $K$, subpopulations $L$, constraints $M$ and weights $\mathbf{q}$. In each round $t=1,2,\cdots$, the algorithm first computes the empirical means of all arms and all subpopulations, denoted by $\hat{\fbseries\mu}(t)\in\mathbb{R}^{KL}$, which is given by $\hat{\mu}_{k,l}(t)=\frac{1}{N_{k,l}(t)}\sum_{s=1}^tX_s\mathbbm{1}(I_s = l, A_s = k).$ Then the algorithm computes a maximizer $\mathbf{w}_t\in\mathbf{w}^*(\hat{\fbseries\mu}(t))$ of problem \eqref{eq:T_star} with ${\fbseries\mu}$ replaced by $\hat{\fbseries\mu}(t)$, which can further be simplified to \eqref{eq:k=0} or \eqref{eq:k=1}. Here, since the maximizer may not be unique, so $\mathbf{w}^*(\hat{\fbseries\mu}(t))$ is defined as the set consists of all maximizers, and we can take $\mathbf{w}_t$ to be any element in $\mathbf{w}^*(\hat{\fbseries\mu}(t))$. Now, we use the C-tracking rule proposed by \cite{garivier2016optimal}. To be specific, if $\varepsilon\in(0,\frac{1}{KL}]$, let $\mathbf{w}^{(\varepsilon)}_t$ be a $L^{\infty}$ projection of $\mathbf{w}_t$ onto $\Sigma_{K\times L}^{\varepsilon}=\{(w_1,\cdots,w_{KL})\in[\varepsilon,1]\big\vert w_1+\cdots+w_{KL}=1\}$, otherwise set $\mathbf{w}^{(\varepsilon)}_t=\mathbf{w}_t$. Take $\varepsilon_t=\left(K^2L^2+t\right)^{-\frac{1}{2}}/2$ and
\begin{equation}\label{eq:c_tracking}
\left(A_{t+1},I_{t+1}\right)\in\arg\max_{(k,l)}\sum_{s=0}^tw_{k,l}^{\varepsilon_s}(\hat{\fbseries\mu}(s))-N_{k,l}(t).    
\end{equation}
Later we will see, this sampling rule ensures that $N_{k,l}(t)$ is close to $\sum_{s=0}^tw_{k,l}^{\varepsilon_s}(\hat{\fbseries\mu}(s))$ and thus close to $tw^*_{k,l}({\fbseries\mu})$, so it is asymptotically optimal and can achieve the lower bound given by \eqref{eq:lower_bound}. 

Now we talk about the calculation of our sampling rule. From above we can see the only challenging aspect is calculating $\mathbf{w}_t$, which is a good approximation of the maximizer of problem \eqref{eq:T_star} with ${\fbseries\mu}$ replaced by $\hat{\fbseries\mu}(t)$. By Theorem \ref{thm:trade_off} we only need to solve the optimization problems 
\eqref{eq:k=0} and \eqref{eq:k=1} with some given ${\fbseries\mu}$. For \eqref{eq:k=0}, it is not hard to see we would expect $\sum_{l\in[M],\mu_{k,l}<0}(\mathbf{w}_t)_{k,l}\mu_{k,l}^2$ to be the same for $k\in [K]$. Define $l(k)=\arg\max_{l\in[M],\mu_{k,l}<0}\mu_{k,l}^2$ and break the tie arbitrarily, then under the constraint $\sum_{k\in[K],l\in[L]}w_{k,l}=1$ we can see $$(\mathbf{w}_t)_{i,l(i)}=\frac{1}{\mu_{i,l(i)}^{2}}\left(\sum_{k\in[K]}\frac{1}{\mu_{k,l(k)}^{2}}\right)^{-1}$$ for $i\in[K]$ and $(\mathbf{w}_t)_{k,l}=0$ for $l\neq l(k)$. Thus, \eqref{eq:k=0} can be explicitly solved. The optimization problem in \eqref{eq:k=1} is more difficult. We first define $F_{\fbseries\mu}(\mathbf{w})=\min\left(f^\text{opt}_{\fbseries\mu}(\mathbf{w}),f^\text{fea}_{\fbseries\mu}(\mathbf{w})\right)$ and calculate $F(\mathbf{w})$ for fixed $\mathbf{w}$ as an optimization problem. Given $\mathbf{w}$, $f^\text{fea}_{\fbseries\mu}(\mathbf{w})=\min_{l\in[M]}w_{1,l}\mu_{1,l}^2$ is known, so it suffices to calculate $f^\text{opt}_{\fbseries\mu}(\mathbf{w})$. Recall that
\[
f^\text{opt}_{\fbseries\mu}(\mathbf{w})=\min_{2\leq k\leq K}\min_{\substack{{\fbseries\lambda}\in \mathbb{R}^{K\times L}\\ \lambda_k\geq \lambda_1\\\forall l\in[M],\lambda_{k,l}\geq0}}\left(\sum_{l\in[L]}w_{1,l}(\mu_{1,l}-\lambda_{1,l})^2+\sum_{l\in[L]}w_{k,l}(\mu_{k,l}-\lambda_{k,l})^2\right),
\]
and for fixed $\mathbf{w}$ and each $k$, the internal minimization programming is a convex quadratic problem with linear constraints, so it can be easily solved through standard optimization methods, say the Lagrangian multiplier method used in Lemma 5 of \cite{russac2021b}. Thus $f^\text{fea}_{\fbseries\mu}(\mathbf{w})$ can be calculated through solving $K-1$ optimization subproblems. 

Now that $F_{\fbseries\mu}(\mathbf{w})$ is known, and it is the minimum of several linear functions of $\mathbf{w}$, so it is concave. In addition, from the definition of $F_{\fbseries\mu}(\mathbf{w})$ we know there exist ${\fbseries\lambda},k\in[K]$ such that $\sum_{l\in[L]}w_{1,l}(\mu_{1,l}-\lambda_{1,l})^2+\sum_{l\in[L]}w_{k,l}(\mu_{k,l}-\lambda_{k,l})^2=F_{\fbseries\mu}(\mathbf{w})$ or there exists $L\in[M]$ such that $w_{1,l}\mu_{1,l}^2=F_{\fbseries\mu}({\mathbf{w}})$. In both cases we can write $F_{\fbseries\mu}(\mathbf{w})=\mathbf{c}^T{\mathbf{w}}$ for some $\mathbf{c}\in\mathbb{R}^{K L}$, so we can obtain a subgradient of $F_{\fbseries\mu}(\mathbf{w})$ given by $\frac{\partial}{\partial{\mathbf{w}}}F_{\fbseries\mu}(\mathbf{w})=\mathbf{c}$. Now, by performing projected subgradient method, we can solve the minimization problem $\max_{\mathbf{w}\in\Sigma_{K\times L}}F(\mathbf{w})$ by updating $w^{(n+1)}=\mathbf{P}_{\Sigma_{K\times L}}\left(\mathbf{w}^{(n)}+\alpha_{n}\frac{\partial}{\partial{\mathbf{w}}}F_{\fbseries\mu}(\mathbf{w}^{(n)})\right)$ iteratively with the projection operator $\mathbf{P}_{\Sigma_{K\times L}}(\cdot)$ and some proper stepsizes $\{\alpha_n\}$, and it is known the projected subgradient method converges under mild conditions, see for example \cite{boyd2003subgradient}. This finishes the calculation of $\mathbf{w}_t$ and also our sampling rule.

\subsection{The stopping rule and the threshold}\label{subsec:stopping_rule}
Following the idea of \cite{garivier2016optimal} and \cite{russac2021b}, we consider the Chernoff's Generalized Likelihood Ratio statistic:
\begin{equation}
\label{eq:stopping_statistics}
Z(t)=\frac{1}{2}\inf_{{\fbseries\lambda}\in \text{Alt}({\fbseries\mu})}\sum_{k\in[K]}\sum_{l\in[L]}N_{k,l}(t)\left(\hat\mu_{k,l}(t)-\lambda_{k,l}\right)^2.
\end{equation}
Note if we define the empirical sampling weights $(\hat{\mathbf{w}}_t)_{k,l}=\frac{N_{k,l}(t)}{t}$, then $Z(t)$ can be written as $
Z(t)=\frac{t}{2}F_{\hat{\fbseries\mu}(t)}(\hat{\mathbf{w}}_t),$ which can be efficiently calculated as dicussed in Section \ref{subsec:sampling_rule}. For a given risk level $0<\delta<1$, we define the stopping time $\tau_\delta$ as follows:
\[
\tau_\delta = \inf_{t\in \mathbb{N}} \{Z(t)>\beta(t,\delta)\}.
\]
Here the threshold $\beta(t,\delta)$ should be tuned appropriately. By Proposition 21 of \cite{kaufmann2021mixture}, a choice of $\beta(t,\delta)=O\left(L\ln\ln t+\log\frac{K}{\delta}\right)$ would ensure our policy to be $\delta$-PAC, while in practice, as suggested by \cite{garivier2016optimal} and \cite{russac2021b}, we use instead the stylized
$\ln({(1+\ln t)}/{\delta})$ which is less conservative. The final recommendation $\hat{k}_{\tau_\delta}$ is just the optimal feasible arm in $\hat{\fbseries\mu}(\tau_\delta)$, i.e. 
$$    \hat{k}_{\tau_\delta}=\arg\max_{k\in{C(\hat{\fbseries\mu}(\tau_\delta))}}\hat{\mu}_{k}.$$
Again, we take $\hat{k}_{\tau_\delta}=0$ if $C(\hat{\fbseries\mu}(\tau_{\delta}))=\emptyset$.

\subsection{The convergence result}\label{subsec:converge}
We now give the convergence result of our algorithm, which matches the asymptotic optimal lower bound given by \eqref{eq:lower_bound}:

\begin{theorem}\label{thm:opt_convergnce}
For every bandit model ${\fbseries\mu}\in\mathcal{S}$, our algorithm is $\delta$-PAC and
\begin{equation}
    \lim_{\delta\to 0}\frac{\mathbb{E}_{\fbseries\mu}[\tau_\delta]}{\ln(1/\delta)}=T^*({\fbseries\mu}).
\end{equation}
\end{theorem}
The proof of Theorem \ref{thm:opt_convergnce} is given as follows. By applying Lemma 7 of \cite{garivier2016optimal} to our C-Tracking rule with $K\times L$ weights, we have
\begin{equation}\label{eq:weight_converge}
\max_{k\in[K],l\in[L]}\big\vert N_{k,l}(t)-\sum_{s=0}^{t-1}\mathbf{w}_t\big\vert\leq KL(1+\sqrt{t}).
\end{equation}
With force exploration rate $\varepsilon_t$, since $t\varepsilon_t=O(\sqrt{t})$, each subpopulation of each arm would be sampled infinite times as $t\to\infty$, so $\hat{\fbseries\mu}(t)\to{\fbseries\mu}$ almost surely. In addition, since $F_{\fbseries\mu}(\mathbf{w})$ is the minimum of several linear functions and thus concave, so the set of maximizers $\mathbf{w}^*({\fbseries\mu})$ is convex, then by Lemma 6 of \cite{degenne2019pure} we know $\inf_{\mathbf{w}\in \mathbf{w}^*({\fbseries\mu})}\Vert\frac{1}{t}\sum_{s=0}^{t-1}\mathbf{w}_t-\mathbf{w}\Vert_\infty\to 0$ as $t\to\infty.$ Combine this with \eqref{eq:weight_converge} we know that $\inf_{\mathbf{w}\in \mathbf{w}^*({\fbseries\mu})}\big\Vert\hat{\mathbf{w}}_t-\mathbf{w}\big\Vert_\infty\to 0$ almost surely, that is, our empirical weights $\hat{\mathbf{w}}_t$ gets close to some oracle weights. Since ${\fbseries\mu}\in\mathcal{S}$ so the problem is single-answered, so by Theorem 7 of \cite{degenne2019pure} we know our algorithm has asymptotically optimal complexity, i.e. $\lim_{\delta\to 0}\frac{\mathbb{E}[\tau_\delta]}{\ln(1/\delta)}=T^*({\fbseries\mu}).$ In addition, by our choice of $\beta(t,\delta)$ and Proposition 21 of \cite{kaufmann2021mixture}, our algorithm is $\delta$-PAC.

\section{Numerical Experiments}
In this section, we demonstrate the efficiency of the Track-and-Stop with fairness Constraints on Subpopulations (T-a-SCS) strategy for addressing BAICS problems through two examples. In the first example, the arm $k$ of maximum quality is infeasible on one subpopulation $l$. More specifically, $\mu_{k,l}<0$ but is close to $0$. The second example presents a situation where two arms have  maximum quality, but one of them is infeasible, further, there is a third arm which is feasible and has a quality close to the maximum quality. Through these examples, we demonstrate the behavior of the algorithm where there is a tradeoff between testing for optimality and testing for feasibility. In comparison with the T-a-SCS strategy, we consider two other benchmark sampling strategies. The first is the original Track-and-Stop (T-a-S) strategy (\cite{garivier2016optimal}), which does not incorporate subpopulation constraints when calculating the weight assignment for each arm. When this strategy is performed, it yields arm $A_t$ to be sampled at iteration $t$. We then randomly allocate the sample to subpopulation $I_t$ of arm $A_t$ with probability $q_{I_t}/\sum_{l\in[L]}q_l$. The second is the uniform sampling strategy. In each iteration, we sample arm $k(t)$ with probability $1/K$, and randomly allocate the sample to subpopulation $I_t$ of arm $A_t$ with probability $q_{I_t}/\sum_{l\in[L]}q_l$. The choice of sampling strategy does not affect the stopping rule. In our experiment, all three algorithms use the Chernoff's Generalized Likelihood Ratio statistic $Z(t)$ given by (\ref{eq:stopping_statistics}), and the same threshold $\beta(t,\delta)=\ln({(1+\ln t)}/{\delta})$.

\subsection{The first example}
\label{subsec:numerical_case1}
In the first numerical case, we set the number of arms and subpopulations, and the respective arm values on each subpopulation as follows:
let $K=3$, $L=3$, ${\fbseries\mu}_1 = (0.2,0.6,0.8)$, ${\fbseries\mu}_2=(0.4,0.4,0.3)$, ${\fbseries\mu}_3=(-0.2,1,1.5)$,  we have noise level $\sigma=1$. In calculating the overall quality of an arm, we set for the three subpopulations $q_1=0.2$, $q_2=0.3$, $q_3=0.5$, and $\mu_k=\sum_{l=1}^3q_k\mu_{k,l}$. In this case, we have $\mu_1=0.62$, $\mu_2=0.35$, $\mu_3=1.01$, but arm $3$ is infeasible because $\mu_{31}<0$, and arm 1 is the best feasible arm. The probability threshold of correct selection is set as $\delta=0.1$.

We initialize each arm with $5$ draws on each subpopulation.  For simplicity, we perform projected subgradient method (see Section \ref{subsec:sampling_rule}) to update $\mathbf{w}$ one time with stepsize $\alpha=1$ in each iteration. The optimal weights $\mathbf{w}_t=(w_1,\cdots,w_K)$ for implementing the T-a-S strategy are calculated by solving a rational equation; we refer to the Gaussian case of \cite{garivier2016optimal} for details. We also use the C-tracking rule for projecting the T-a-S weights.
We run 300 experiments to record the average stopping time $\hat{\tau}_\delta$ and empirical probability of correct selection $\hat{P}_{\fbseries\mu}(\hat{k}_{\tau_\delta}=k^*({\fbseries\mu}))$. The results are given in table \ref{table:case1}.

\begin{table}[ht]
\centering
\begin{tabular}{c|c|c|c}
  \hline
  &T-a-SCS & T-a-S & Uniform \\
  \hline
 $\hat{\tau}_\delta$ &530 & 1703 & 2432 \\
  \hline
$\hat{P}_{\fbseries\mu}$& 0.987 & 0.990 & 0.983\\
  \hline
  
\end{tabular}
\caption{Average stopping time and empirical probability of correct selection of the three sampling strategies in example 1.}
\label{table:case1}
\end{table}

\begin{figure}[htbp]
  \centering
  \subfigure[T-a-SCS]{
  \begin{minipage}{0.48\textwidth}
    \centering
    \includegraphics[width=\linewidth]{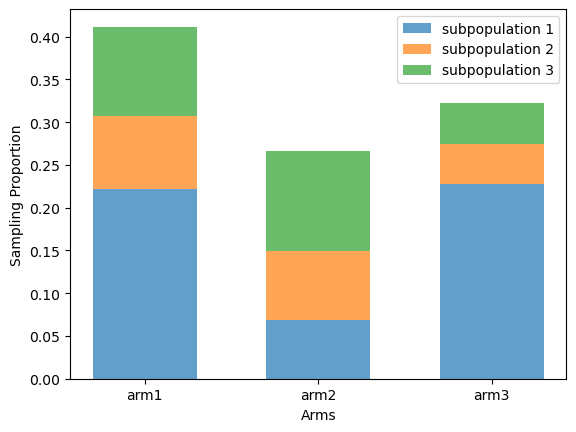}
    \label{fig:sub1}
  \end{minipage}}
  \subfigure[T-a-S]{
  \begin{minipage}{0.48\textwidth}
    \centering\includegraphics[width=\linewidth]{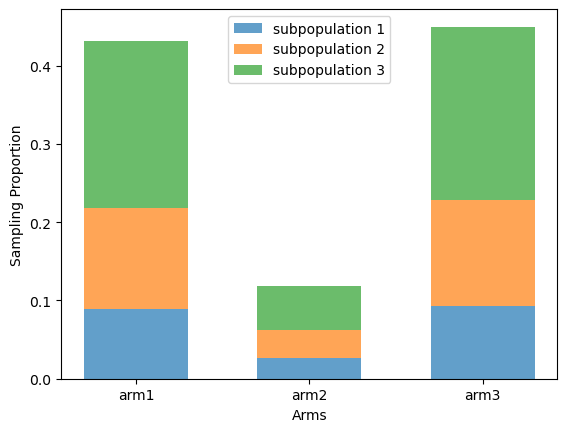}
    \label{fig:sub2}
  \end{minipage}}
  \caption{Average sample allocation to the arms and the subpopulations in example 1.}
  \label{fig:case1}
\end{figure}

We further look at how the samples are allocated to each of the arms and subpopulations in the T-a-SCS strategy, in comparison with the T-a-S strategy. For each experiment, we record the number of samples on each arm and subpopulation, $\{N_{k,l}(\tau_\delta):k\in[K],l\in[L]\}$, and compute the empirical sampling weights $(\hat{\mathbf{w}})_{k,l}=\frac{N_{k,l}(\tau_\delta)}{\tau_\delta}$. We then take an average over the 300 copies of experiments. The results are given in Figure \ref{fig:case1}, demonstrating a tradeoff between  optimality and feasibility. Compared to the T-a-S strategy, we notice that the T-a-SCS strategy assigns more empirical sampling weights to the subpopulations on which the arm values are close to $0$ (e.g., subpopulation 1 of arm 1, subpopulation 1 of arm 3). Further, as the infeasibility of arm 3 is ``discovered" by T-a-SCS, it allocates more samples to arm 2 (compared to the T-a-S strategy), which is now the only competitor for arm 1 of being the best feasible arm.

\subsection{The second example}
In the second numerical case, we set the number of arms and subpopulations, and the respective arm values on each subpopulation as follows:
let $K=4$, $L=3$, ${\fbseries\mu}_1 = (-0.2,0.4,1.2)$, ${\fbseries\mu}_2=(0.2,0.6,0.6)$, ${\fbseries\mu}_3=(0.3,0.3,0.6)$, ${\fbseries\mu}_4=(-0.6,0.8,0.4)$, we have noise level $\sigma=1$. In calculating the overall quality of an arm, we set for the three subpopulations $q_1=q_2=q_3=1/3$, and ${\mu}_k=\sum_{l=1}^3q_k\mu_{k,l}$. In this example we have ${\mu}_1={\mu}_2=0.47$, which equals the maximum quality, but arm 1 is infeasible. Also, ${\mu}_3$ is $0.4$, which is close to ${\mu}_2$. The probability threshold of correct selection is set as $\delta=0.1$.

The numerical settings are similar to that of Section \ref{subsec:numerical_case1}. In this example, both the T-a-S and the Uniform sampling strategy exceed the limit of $\tau_{\max}=15000$ iterations in a large proportion of experiment copies. For T-a-SCS we have $\hat{\tau}_\delta=3131$ and $\hat{P}_{\fbseries\mu}(\hat{k}_{\tau_\delta}=k^*({\fbseries\mu}))=0.980$. 

\begin{figure}[htbp]
  \centering
  \subfigure[T-a-SCS]{
  \begin{minipage}{0.48\textwidth}
    \centering
    \includegraphics[width=\linewidth]{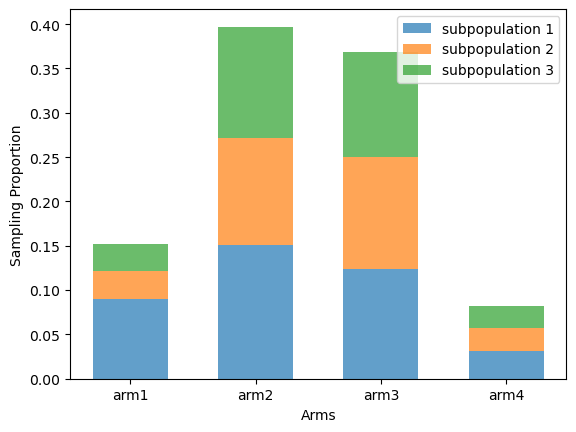}
    \label{fig:sub12}
  \end{minipage}}
 \subfigure[T-a-S]{
  \begin{minipage}{0.48\textwidth}
    \centering
    \includegraphics[width=\linewidth]{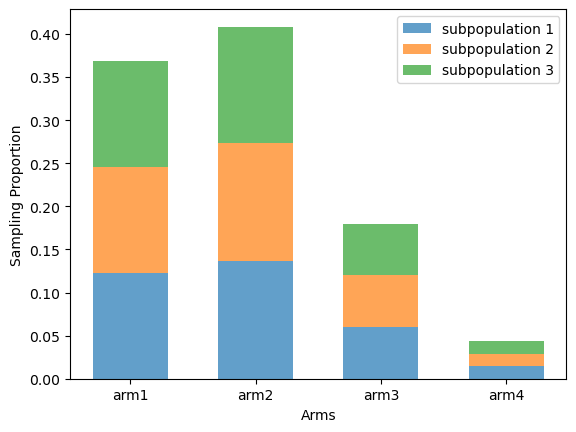}
    \label{fig:sub22}
  \end{minipage}}
  \caption{Average sample allocation to the arms and the subpopulations in example 2.}
  \label{fig:case2}
\end{figure}

We further look at how the samples are allocated to each of the arms and subpopulations in the T-a-SCS strategy, in comparison with the T-a-S strategy. The results are given in figure \ref{fig:case2}. In this example, although arm 1 has the maximum quality, the T-a-SCS strategy ``realizes" that arm 1 is very likely infeasible because of the negative values on subpopulation 1. The T-a-SCS strategy thus assigns less empirical sampling weight to arm 1 (in comparison to the T-a-S strategy) aside from checking its feasibility on subpopulation 1. This further allows the T-a-SCS strategy to assign more empirical weight to the other feasible arm 3, and to arrive at the conclusion that arm 2 has better quality than arm 3, and is therefore the best feasible arm, with less sampling times. 

\section{Conclusion}
We formulate, analyze and solve the problem of best arm identification with fairness constraints on subpopulations (BAICS). The BAICS problem requires that an selected arm must be fair to all subpopulations by satisfying constraints to regulate conditional expected rewards on each subpopulation. The BAICS problem aims at correctly identify the best arm among all feasible arms. We analyze the complexity of the BAICS problem by proving a best achievable lower bound on the sample complexity. We then design an algorithm, and prove the sample complexity to match with the lower bound in terms of order. A brief account of numerical experiments are conducted to illustrate the theoretical findings. 

%% file: arXiv_version_230407.bbl
\begin{thebibliography}{20}
\providecommand{\natexlab}[1]{#1}
\providecommand{\url}[1]{\texttt{#1}}
\providecommand{\urlprefix}{URL }

\bibitem[{Andrad{\'o}ttir \protect\BIBand{} Kim(2010)}]{andradottir2010fully}
Andrad{\'o}ttir S, Kim SH (2010) Fully sequential procedures for comparing
  constrained systems via simulation. \emph{Naval Research Logistics (NRL)}
  57(5):403--421.

\bibitem[{Audibert et~al.(2010)Audibert, Bubeck, \protect\BIBand{}
  Munos}]{audibert2010best}
Audibert JY, Bubeck S, Munos R (2010) Best arm identification in multi-armed
  bandits. \emph{COLT}, 41--53.

\bibitem[{Barrier et~al.(2022)Barrier, Garivier, \protect\BIBand{}
  Koc{\'a}k}]{barrier2022non}
Barrier A, Garivier A, Koc{\'a}k T (2022) A non-asymptotic approach to best-arm
  identification for gaussian bandits. \emph{International Conference on
  Artificial Intelligence and Statistics}, 10078--10109 (PMLR).

\bibitem[{Bechhofer et~al.(1995)Bechhofer, Santner, \protect\BIBand{}
  Goldsman}]{bechhofer1995design}
Bechhofer R, Santner T, Goldsman D (1995) Design and analysis of experiments
  for statistical selection, screening, and multiple comparison john wiley and
  sons. \emph{Hoboken, New Jersey} .

\bibitem[{Bechhofer(1954)}]{bechhofer1954single}
Bechhofer RE (1954) A single-sample multiple decision procedure for ranking
  means of normal populations with known variances. \emph{The Annals of
  Mathematical Statistics} 16--39.

\bibitem[{Boyd et~al.(2003)Boyd, Xiao, \protect\BIBand{}
  Mutapcic}]{boyd2003subgradient}
Boyd S, Xiao L, Mutapcic A (2003) Subgradient methods .

\bibitem[{Chick(2006)}]{chick2006subjective}
Chick SE (2006) Subjective probability and bayesian methodology.
  \emph{Handbooks in Operations Research and Management Science} 13:225--257.

\bibitem[{Degenne \protect\BIBand{} Koolen(2019)}]{degenne2019pure}
Degenne R, Koolen WM (2019) Pure exploration with multiple correct answers.
  \emph{Advances in Neural Information Processing Systems} 32.

\bibitem[{Garivier \protect\BIBand{} Kaufmann(2016)}]{garivier2016optimal}
Garivier A, Kaufmann E (2016) Optimal best arm identification with fixed
  confidence. \emph{Conference on Learning Theory}, 998--1027 (PMLR).

\bibitem[{Glynn \protect\BIBand{} Juneja(2015)}]{glynn2015selecting}
Glynn P, Juneja S (2015) Selecting the best system and multi-armed bandits.
  \emph{arXiv preprint arXiv:1507.04564} .

\bibitem[{Healey et~al.(2014)Healey, Andrad{\'o}ttir, \protect\BIBand{}
  Kim}]{healey2014selection}
Healey C, Andrad{\'o}ttir S, Kim SH (2014) Selection procedures for simulations
  with multiple constraints under independent and correlated sampling.
  \emph{ACM Transactions on Modeling and Computer Simulation (TOMACS)}
  24(3):1--25.

\bibitem[{Hong et~al.(2021)Hong, Fan, \protect\BIBand{} Luo}]{hong2021review}
Hong LJ, Fan W, Luo J (2021) Review on ranking and selection: A new
  perspective. \emph{Frontiers of Engineering Management} 8(3):321--343.

\bibitem[{Hong et~al.(2015)Hong, Luo, \protect\BIBand{}
  Nelson}]{hong2015chance}
Hong LJ, Luo J, Nelson BL (2015) Chance constrained selection of the best.
  \emph{INFORMS Journal on Computing} 27(2):317--334.

\bibitem[{Hunter \protect\BIBand{} Nelson(2017)}]{hunter2017parallel}
Hunter SR, Nelson BL (2017) Parallel ranking and selection. \emph{Advances in
  Modeling and Simulation}, 249--275 (Springer).

\bibitem[{Kaufmann et~al.(2016)Kaufmann, Capp{\'e}, \protect\BIBand{}
  Garivier}]{kaufmann2016complexity}
Kaufmann E, Capp{\'e} O, Garivier A (2016) On the complexity of best arm
  identification in multi-armed bandit models. \emph{Journal of Machine
  Learning Research} 17:1--42.

\bibitem[{Kaufmann \protect\BIBand{} Koolen(2021)}]{kaufmann2021mixture}
Kaufmann E, Koolen WM (2021) Mixture martingales revisited with applications to
  sequential tests and confidence intervals. \emph{The Journal of Machine
  Learning Research} 22(1):11140--11183.

\bibitem[{Kim \protect\BIBand{} Nelson(2006)}]{kim2006selecting}
Kim SH, Nelson BL (2006) Selecting the best system. \emph{Handbooks in
  operations research and management science} 13:501--534.

\bibitem[{Ma \protect\BIBand{} Henderson(2017)}]{ma2017efficient}
Ma S, Henderson SG (2017) An efficient fully sequential selection procedure
  guaranteeing probably approximately correct selection. \emph{2017 Winter
  Simulation Conference (WSC)}, 2225--2236 (IEEE).

\bibitem[{Russac et~al.(2021)Russac, Katsimerou, Bohle, Capp{\'e}, Garivier,
  \protect\BIBand{} Koolen}]{russac2021b}
Russac Y, Katsimerou C, Bohle D, Capp{\'e} O, Garivier A, Koolen WM (2021)
  A/b/n testing with control in the presence of subpopulations. \emph{Advances
  in Neural Information Processing Systems} 34:25100--25110.

\bibitem[{Shang et~al.(2020)Shang, Heide, Menard, Kaufmann, \protect\BIBand{}
  Valko}]{shang2020fixed}
Shang X, Heide R, Menard P, Kaufmann E, Valko M (2020) Fixed-confidence
  guarantees for bayesian best-arm identification. \emph{International
  Conference on Artificial Intelligence and Statistics}, 1823--1832 (PMLR).

\end{thebibliography}
